\pdfoutput=1

\documentclass[11pt]{article}

\usepackage[]{acl}

\usepackage{times}
\usepackage{latexsym}
\usepackage{amsmath}
\usepackage[T1]{fontenc}
\usepackage{graphicx}
\usepackage[utf8]{inputenc}

\usepackage{microtype}

\title{BLINK with Elasticsearch for Efficient Entity Linking in \\ Business Conversations}

\author{Md Tahmid Rahman Laskar, Cheng Chen, Aliaksandr Martsinovich,  Jonathan Johnston, \\ \textbf{Xue-Yong Fu}, \textbf{Shashi Bhushan TN}, \textbf{Simon Corston-Oliver} \\
          Dialpad Canada Inc. \\
    1100 Melville St \#400 \\ Vancouver, BC, Canada, V6E 4A6 \\
  \texttt{\{tahmid.rahman,cchen,aliaksandr.martsinovich,jonathan\}@dialpad.com}\\
  \texttt{\{xue-yong,sbhushan,scorston-oliver\}@dialpad.com}}

\begin{document}
\maketitle
\begin{abstract}
An Entity Linking system aligns the textual mentions of entities in a text to their corresponding entries in a knowledge base. However, deploying a neural entity linking system for efficient real-time inference in production environments is a challenging task. In this work, we present a neural entity linking system that connects the product and organization type entities in business conversations to their corresponding Wikipedia and Wikidata entries. 
The proposed system leverages Elasticsearch to ensure inference efficiency when deployed in a resource limited cloud machine, and obtains significant improvements in terms of inference speed and memory consumption while retaining high accuracy. 

\end{abstract}


\section{Introduction}

Companies that offer VoIP telephony products with built-in speech and natural language processing features aim to assist the customer support agents with information relevant to the content of their conversations with the customers. To be useful, such assistance should be provided in near real-time of the triggering utterance. 
In this paper, we demonstrate how we build a near real-time entity linking system at Dialpad\footnote{\url{https://www.dialpad.com/}} to link the entities in business phone transcripts to a knowledge base to provide more semantically-informed assistance.

The entity linking task is usually comprised of three steps: (i) detect the mentions in the given text, (ii) generate a list of candidate entities relevant to each mention, and finally (iii) link each mention to its most relevant entry in the knowledge base \cite{DBLP:conf/eacl/cholan}. Note that entity linking systems used in production should provide the optimum performance in terms of both inference speed and memory consumption while being used within a limited computational budget. Since there are millions of entities stored in a knowledge base, the scaling issue is a major concern while developing a real-time entity linking system. 

The goal of this research is to develop a neural entity linking system to efficiently link \textit{product} and \textit{organization} type entities in business phone conversations to their respective entries in a knowledge base for information extraction. For that purpose, we present an extended version of the state-of-the-art neural entity linker, the BLINK model \cite{DBLP:conf/emnlp/blink}. Though BLINK was originally proposed for entity linking on Wikipedia, we extend it for entity linking on Wikidata\footnote{\url{https://www.wikidata.org/}} since unlike Wikipedia, the Wikidata knowledge base contains information related to the entities in a structured way. Thus, it allows effective extraction of relevant information for each entity. More importantly, for production deployment, we also introduce several new techniques that significantly reduce the memory requirements, computational resource usage, and the inference speed of BLINK. More concretely, our major contributions are stated below:
 
 \begin{figure*}[t]
\centering
\includegraphics[width=\linewidth]{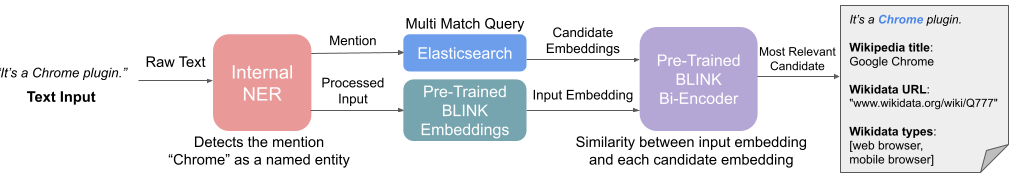} 
\caption{The Proposed Entity Linking System. First, our Internal NER model detects the mention in the given text. Then we retrieve a list of candidate entities with their embeddings from Elasticsearch. At the same time, we generate the contextualized representation of the input text using the pre-trained BLINK embeddings. Afterward, we utilize the pre-trained BLINK Bi-Encoder to determine the entity that is the most relevant among the candidates and finally we extract information related to that entity from our knowledge base in Elasticsearch.}
\label{fig1}
\end{figure*}

\begin{itemize}
    \item We tackle the computational complexities in BLINK by saving all pre-trained entity embeddings in Elasticsearch\footnote{\url{https://www.elastic.co/elasticsearch/}} and propose a word matching technique to retrieve the candidate entities faster. We also present an approach to pre-compute the linking between the Wikipedia page of each entity to its respective Wikidata page to reduce the runtime latency.
    \item Extensive experiments show that our entity linking system significantly reduces the inference time and memory requirements while retaining high accuracy in a computationally inexpensive machine. We also successfully deploy our entity linking system in a 10GB RAM machine (without GPU) whereas the original model requires a machine in our server having 60 GB RAM for inference.
    
\end{itemize}

\section{Related Work}

Prior work on entity linking mostly focused on linking named entities to unstructured knowledge bases like Wikipedia, whereas the amount of work that used a structured knowledge base like Wikidata is very limited \cite{shen2014entity,DBLP:conf/cikm/falcon}. Though other knowledge bases like DBpedia \cite{auer2007dbpedia} or YAGO \cite{fabian2007yago} have also been studied, the utilization of Wikidata as the knowledge base to extract relevant information has gained lots of attention recently \cite{lin2021rockner,moller2021survey}.  

Detecting mentions (i.e., entities) in the given text \cite{DBLP:journals/corr/HuangXY15,DBLP:conf/coling/AkbikBV18} is an important step for entity linking. In recent years, utilizing the neural network architecture for mention detection has been extensively studied \cite{DBLP:conf/emnlp/blink,DBLP:conf/aaai/OnoeD20}. More recently, the impressive success of the transformer architecture \cite{DBLP:conf/nips/VaswaniSPUJGKP17,DBLP:conf/naacl/bert, DBLP:conf/emnlp/YamadaASTM20} in a wide range of natural language processing tasks has also inspired researchers to apply transformer models for the entity recognition \cite{lin2021rockner} step in entity linking \cite{DBLP:conf/eacl/cholan}, which results in obtaining superior performance over the previously used recurrent neural network-based models \cite{peters-etal-2018-deep}. 

For the candidate generation step in entity linking, early work mostly utilized various non-neural network approaches such as TF-IDF or alias tables \cite{DBLP:conf/emnlp/blink}, whereas more recent work utilized dense embeddings learnt via pre-trained transformers to retrieve the relevant candidates \cite{DBLP:conf/emnlp/blink,onoe2020interpretable}. However, there is an important limitation while generating the candidates via pre-trained embeddings. For instance, the state-of-the-art neural entity linking model BLINK \cite{DBLP:conf/emnlp/blink} loads the pre-trained embeddings of all entities in Wikipedia into memory. Thus, it becomes inapplicable for deployment in production scenarios where the requirement is to ensure lower memory consumption. In this paper, we address this issue via storing the pre-trained embeddings in Elasticsearch. Moreover, we introduce new techniques that pre-compute the linking between Wikipedia and Wikidata to ensure efficient information retrieval, while also optimize the pre-trained models to meet the goal of deploying the proposed system in a limited computational resource setting.

\section{System Overview}
To develop the entity linking system, we adopt BLINK, a neural entity linker that uses the transformer-based BERT model \cite{DBLP:conf/nips/VaswaniSPUJGKP17,DBLP:conf/naacl/bert} and trains it on Wikipedia. BLINK connects each mention in a given text with its respective Wikipedia page based on the overall context. Since Wikipedia contains textual data in an unstructured format, it is difficult to extract information from it. Thus, we connect BLINK with a structured knowledge base, Wikidata, to extract information about product and organization type entities. 
Note that we store our knowledge base as well as the embedding representation of each entity in Elasticsearch. Moreover, we replace the Flair Named Entity Recognition (NER) model \cite{DBLP:conf/naacl/AkbikBBRSV19} originally used by BLINK with an NER model (we denote it as \textbf{Internal NER}) trained on transcripts of business phone conversations using DistilBERT \cite{DBLP:journals/corr/distilbert}. 

We show our entity linking system in Figure~\ref{fig1}. At first, the input text is processed by the NER model to detect the mention. Then, we generate the representation for the input text using the pre-trained BLINK embeddings, while we retrieve the relevant candidates with their embeddings from Elasticsearch using the Multi Match Query\footnote{\url{https://www.elastic.co/guide/en/elasticsearch/reference/current/query-dsl-multi-match-query.html}} feature of Elasticsearch. Finally, the embedding representations of the input text and the candidates are sent to the pre-trained BLINK Bi-Encoder to select the most relevant candidate. Below, we first demonstrate our proposed entity linking system: \textit{\textbf{BLINK with Elasticsearch}}, followed by describing how we deploy our proposed system in production. 


\begin{figure}[t]
\centering
\includegraphics[height=6cm]{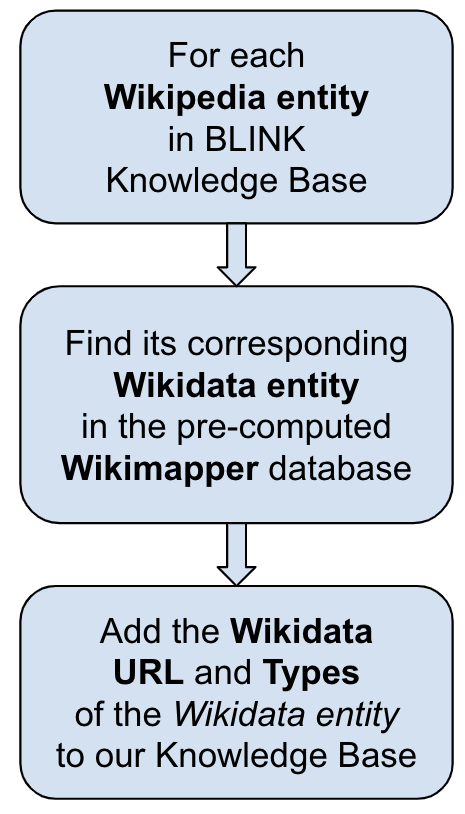} 
\caption{Precomputing Wikipedia to Wikidata Linking.}
\label{fig2}
\end{figure}

\subsection{BLINK with Elasticsearch}

The original BLINK model requires about 25GB RAM to load all pretrained embeddings into memory. In our proposed system, we instead store these embeddings in an external database. To do so, we store all entity embeddings as dense vectors\footnote{\url{https://www.elastic.co/guide/en/elasticsearch/reference/current/dense-vector.html}} in our knowledge base in a remote Elasticsearch server along with saving textual information, such as Wikipedia title, description, URL, and etc. of each entity. This allows the model to only load the top $K$ candidate embeddings into the memory that are most relevant to the mention in a given utterance. As mentioned earlier, the BLINK model was trained over Wikipedia, while our goal is to utilize Wikidata for information extraction. Thus, we need to map the Wikipedia URL of each entity to its Wikidata URL such that we can utilize Wikidata to extract relevant information.  Below, we first describe how we add Wikidata URL of each entity to our knowledge base. Then, we demonstrate how we retrieve the relevant candidates from our knowledge base. 

\subsubsection{Pre-computing Wikipedia to Wikidata Linking} We pre-compute the mapping between Wikipedia and Wikidata using the Wikimapper\footnote{\url{https://github.com/jcklie/wikimapper}} API and add the Wikidata URL of each entity to our knowledge base in Elasticsearch (see Figure \ref{fig2}). This allows our entity linking system to reduce the runtime latency. Note that during the pre-computation step, other information from Wikidata for each entity can also be added to the knowledge base (for our case, we add the \textit{instance of} property as the entity type). 

\subsubsection{Multi Match Query for Candidate Retrieval}
\begin{figure}[t]
\centering
\includegraphics[width = 7cm]{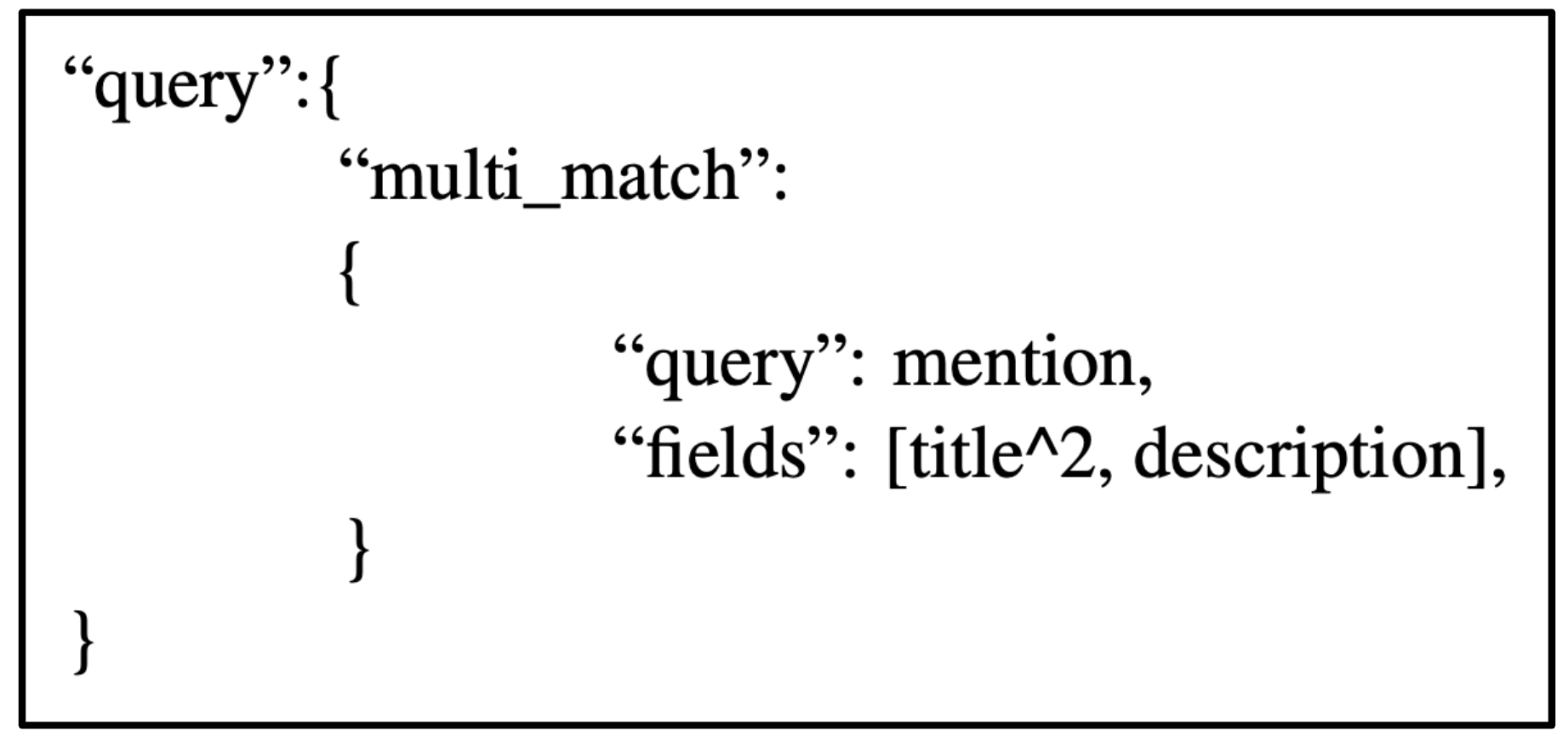} 
\caption{Our Multi Match Query in Elasticsearch.}
\label{figmmq}
\end{figure}

We find that the whole word or subword(s) in the product or organization type entity names usually appear in the Wikipedia \textit{title} and \textit{description} fields. Thus, to retrieve the most relevant candidates, we utilize the \textit{multi match query} feature of Elasticsearch for each entity mention in the input text and apply it to the \textit{title} and \textit{description} fields in our knowledge base (see Figure \ref{figmmq}). For \textit{multi match query}, we give more weight to the \textit{title} field to make it two times more important than the \textit{description} field. In this way, we retrieve the top $k = 250$ candidates from Elasticsearch and send to the BLINK Bi-Encoder to select the most relevant entity.



\subsection{Model Deployment}

\begin{figure*}[t]
\centering
\includegraphics[width=\linewidth]{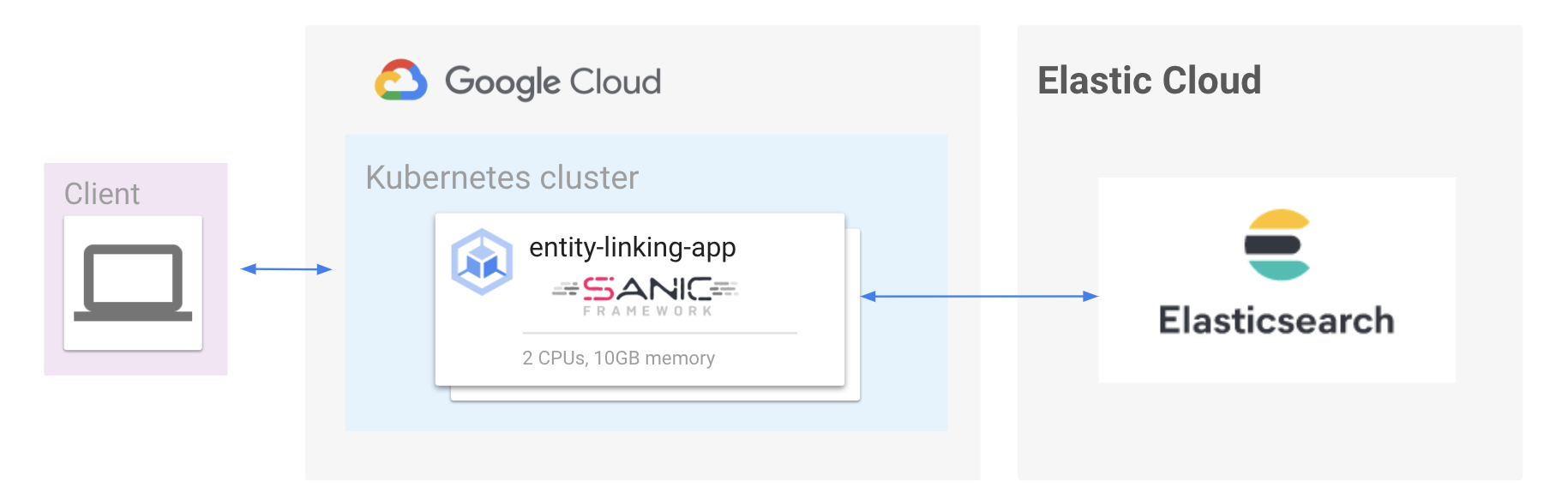} 
\caption{Deployed System Architecture.}
\label{fig3}
\end{figure*}

We deploy our entity linking system in containers\footnote{\url{https://cloud.google.com/kubernetes-engine}} in a Kubernetes\footnote{\url{https://kubernetes.io/}} cluster with 2 CPUs and 10GB RAM. The deployed system architecture is shown in Figure \ref{fig3}. For production deployment, we also apply some optimization techniques to reduce the size of the pre-trained Bi-Encoder, as well as our knowledge base. We describe these below.  

\subsubsection{Pre-trained Bi-Encoder Optimization}

We noticed that the binary file of the pre-trained BLINK Bi-Encoder had two type of tensors: one for context encoding (for the input representation), and the other for the candidate encoding. However, the candidate encoding is only required during the training phase and it is not required during the inference stage since all the candidate embeddings are already stored in our knowledge base in Elasticsearch. Thus, we remove the unnecessary candidate encoding tensors from the binary file which results in reducing the file size from 2.5GB to 1.2GB (50\% reduced space) to improve memory efficiency.

\subsubsection{Knowledge Base Optimization}

The original version of the pre-trained BLINK model \cite{DBLP:conf/emnlp/blink} learns the embedding representations of 59,03,527 Wikipedia entities. In total, the size of these pre-computed embeddings is about 23GB. As our goal is to detect the \textit{Product} and \textit{Organization} type entities in business conversational data, we apply some filtering techniques to optimize the knowledge base such that it mostly contains the entities that are relevant to our NER system. In order to do that, we utilize the \textit{Instance Of} property in Wikidata of each entity and remove entities that are of \textit{Person, Disambiguation, Location, etc}. In this way, the size of the Knowledge base is reduced from 23GB to 12GB (about 50\% reduced space), while the total number of entities has been reduced from 59,03,527 to 27,84,042.

\section{Experimental Details}

In this section, we demonstrate the datasets used in our experiments and the implementation details. 

\subsection{Datasets} To demonstrate the effectiveness of our proposed approach, we conduct a series of experiments on seven academic datasets as well as on a sample of 287 utterances collected from business conversation data. Below, we describe these datasets.

\subsubsection{Business Conversation Dataset}

As our goal is to develop an entity linking system that can link entities in conversational data from  business domains, we sample some real world business phone conversation transcripts. After data collection, we use domain experts (in-house scientists) to annotate the utterances to label the mentions (i.e., product and organization type entities). Our annotated business conversation data consists of 287 utterances that we use in our experiment for evaluation. 

\subsubsection{Academic Datasets}

Since our goal is to develop an entity linking system to extract information for product and organization type entities, at first we pre-process the academic datasets such that our model only links product and organization type entities during experiments. Similar to the original BLINK model \cite{DBLP:conf/emnlp/blink}, we also did not leverage the training data and only used the test data of each dataset for zero-shot entity linking. In our experiment, we use the AIDA-YAGO2-CONLL dataset (testa and testb) from \citet{hoffart2011robust} that contains newswire articles from the Reuters Corpus; the ACE 2004, AQUAINT, and MSNBC datasets from \cite{guo2018robust} that were constructed from news articles; and the WNED-CWEB \cite{guo2018robust} and the WNED-WIKI \cite{wikiwned} datasets that were constructed from CWEB and Wikipedia respectively.

\subsection{Implementation}

Recall that instead of using the Flair NER model \cite{DBLP:conf/naacl/AkbikBBRSV19} used by the original BLINK model, we train an NER model on phone transcripts as our goal is to build the entity linking model for real world business conversation data. For this purpose, we adopt the pre-trained DistilBERT model \cite{DBLP:journals/corr/distilbert} and fine-tune it on a business conversational dataset collected from some phone transcripts in Dialpad that contains 516124 training samples (16124 instances were annotated by humans while 500k instances were pseudo labels generated by the pre-trained LUKE NER model \cite{DBLP:conf/emnlp/YamadaASTM20}). There were also 2292 human annotated samples in the validation set while 4497 human annotated samples in the test set. We use the HuggingFace\footnote{\url{https://github.com/huggingface}} library \cite{wolf2020transformers} to implement the distilbert-base-cased\footnote{\url{https://huggingface.co/distilbert-base-cased/blob/main/config.json}} model and utilize it for the sequence labeling task with the following hyperparameters: \textit{learning rate = 2e-5, total number of epoch = 15,} and \textit{batch size = 32}.  To implement the BLINK model for inference, we use its original source code\footnote{\url{https://github.com/facebookresearch/BLINK}}.

\begin{table*}[t]
\centering
\begin{tabular}{c|c|c|c|c}
    \textbf{Model} & \textbf{NER} & \textbf{Avg. Inf. Time} & \textbf{Accuracy} & \textbf{Memory} \\
    BLINK + PWB & Flair & 2.45 & 62.72 & 	60 GB \\
    BLINK\textsubscript{FAISS} + PWB & Flair & 2.34 & 60.28 & 	60 GB \\
    BLINK + PWB & Internal & 2.79 & 91.64 & 	60 GB \\
    BLINK\textsubscript{FAISS} + PWB & Internal & 2.71 & 88.50 & 	60 GB \\
    BLINK + ES\textsubscript{CS} & Internal & 13.93 & 75.96 & 15 GB \\
    \textbf{BLINK + ES\textsubscript{MMQ}} & \textbf{Internal} & \textbf{1.76} & \textbf{93.03} & \textbf{15 GB} \\
\end{tabular}
\caption{Experimental Results on a sample of 287 utterances. Here, ``Avg. Inf. Time'' refers to ``Average Inference Time in seconds per utterance'', 
``Memory'' refers to the RAM configuration of the Machine that was used. Moreover, we refer the DistilBERT model fine-tuned on phone conversational transcripts as the ``Internal'' NER model.}
\label{table1}
\end{table*}

\begin{table*}[t]
\centering
\begin{tabular}{c|c|c|c}
    \textbf{Datasets} & \textbf{BLINK + ES\textsubscript{CS}}  & \textbf{BLINK + ES\textsubscript{MMQ}} & \textbf{Total Instances} \\
    AIDA-YAGO2-CONLL (testa) & \textbf{67.83} & 62.84 & 3407 \\
    AIDA-YAGO2-CONLL (testb) & 63.74 & \textbf{65.64} & 3425 \\
    ACE 2004  & 75.12 & \textbf{82.95} & 217 \\
    AQUAINT & \textbf{76.96} & 76.46 & 599 \\
    MSNBC & 71.50 & \textbf{79.02} & 386 \\
    WNED-CWEB  & 54.98 & \textbf{61.15} & 8834 \\
    WNED-WIKI & 70.07 & \textbf{74.10} & 5617 \\
\end{tabular}
\caption{Experimental Results on academic datasets based on Cosine Similarity (CS) vs Multi Match Query (MMQ). Here, we use Accuracy as the evaluation metric.}
\label{table2}
\end{table*}

\section{Results and Discussions}

We denote our entity linking model that utilizes Multi Match Query (MMQ) on Elasticsearch (ES) as \textbf{BLINK + ES\textsubscript{MMQ}}. Here, we first discuss its performance on our business conversation data. Then we conduct experiments on some academic datasets to demonstrate its generalized effectiveness. 


\subsection{Performance on Business Conversation Data}

Below, we present some baselines that we use to compare the performance of our proposed model. 

\noindent \textbf{BLINK + PWB:} This model adopts the original BLINK model for entity linking on Wikipedia and utilizes Pywikibot\footnote{\url{https://www.mediawiki.org/wiki/Manual:Pywikibot}}(PWB) for linking between Wikipedia and Wikidata. 

\noindent  \textbf{BLINK\textsubscript{FAISS} + PWB:} This model is similar to the above but utilizes the approximate nearest neighbour search using FAISS \cite{DBLP:journals/tbd/faiss}. 

\noindent \textbf{BLINK + ES\textsubscript{CS}:} This model is similar to our proposed model but uses the Cosine Similarity (CS) feature of Elasticsearch instead of MMQ to retrieve the candidate entities. 

For this experiment, we use the following evaluation metrics, \textbf{(i) average inference time:} \textit{it refers to how much time it takes on average per utterance for entity linking,} {(ii) \textbf{accuracy:}} \textit{it computes the correctness of linking the named entities to the Wikidata knowledge base,} {(iii) \textbf{memory:}} \textit{it refers to the RAM configuration of the Machine that had to be used to run the model in \textit{Google Cloud Platform (GCP)\footnote{\url{https://cloud.google.com/}}} .}

Since the utilization of GPUs significantly increases the computational cost, we did not leverage any GPU in our experiments to mimic the production environment. We show our experimental results in Table \ref{table1} and find that 
our proposed model significantly reduces the inference time while achieving high accuracy. Moreover, we were able to run our proposed model in GCP
on an \textit{n1-standard-4} machine having 15GB RAM with 4 CPUs whereas BLINK models with Pywikibot had to be run on an \textit{n1-standard-16} machine having 60GB RAM with 16 CPUs (we failed to run the model for inference due to memory leaks in other \textit{n1-standard} machines in GCP that had less RAM). 

From Table \ref{table1}, we also observe that the performance of BLINK + ES\textsubscript{CS} model is the poorest among all models. One possible explanation behind this could be because the BLINK model did not leverage cosine similarity during its training phase and so zero-shot cosine similarity between the embedding of the candidate entity and the input embedding for candidate entity retrieval led to poorer accuracy. Moreover, we observe that the cosine similarity between embeddings is also very slow in comparison to MMQ. Furthermore, we find that our Internal NER is more effective than the Flair NER (about 46\%) and combining it with the MMQ leads to the highest accuracy score of $93.03$. 

\subsection{Performance on Academic Datasets}

In this section, we further analyze the performance of our proposed \textbf{BLINK + ES\textsubscript{MMQ}} model via conducting experiments on seven academic datasets. We particularly conduct this experiment to investigate the generalized effectiveness of multi match query. For this analysis, we use the \textbf{BLINK + ES\textsubscript{CS}} model as the baseline where cosine similarity has been used instead of multi match query. As our goal is to deploy our model in a limited computational resource setting to ensure less memory consumption, we only use the models in this experiment that can be run in a machine that do not require more than 16GB RAM. For this reason, we use the models that leverage Elasticsearch instead of Pywikibot (we have already demonstrated in our previous experiment on business conversation data how our proposed method is more effective in terms of both accuracy and efficiency than other baseline models that utilized Pywikbot). 

We show the results of our experiments in Table \ref{table2} to find that in 5 out of 7 datasets, our proposed method that uses multi match query instead of cosine similarity outperforms its counterparts. The only two datasets where our model could not outperform the baseline are the AIDA-YAGO2-CONLL dataset (testa) and the AQUAINT dataset where cosine similarity outperforms multi match query by 7.94\% and 0.65\% respectively. In other datasets, our proposed \textbf{BLINK + ES\textsubscript{MMQ}} model outperforms the \textbf{BLINK + ES\textsubscript{CS}} model by $2.98\%$, $10.42\%$, $10.52\%$, $11.22\%$, and $5.75\%$ in AIDA-YAGO2-CONLL (testb), ACE 2004, MSNBC, WNED-CWEB, and WNED-WIKI datasets respectively. Furthermore, we find during our experiments that our proposed method outperforms its counterpart in terms of inference speed in all 7 datasets (on average, 8 times faster). These findings further validate the effectiveness of our proposed \textbf{BLINK + ES\textsubscript{MMQ}} model for real world deployment in computationally limited resource settings. 

So far, we discuss the effectiveness of our entity linking system in terms of both accuracy and efficiency based on extensive experiments in business conversation data, as well as in benchmark academic datasets. Below, we conduct a case study to analyze how the top $K$ candidates retrieval from Elasticsearch impacts the overall performance.

\begin{table}[t]
\centering
\begin{tabular}{c|c|c}
    \textbf{Top K} & \textbf{Avg. Inf. Time} & \textbf{Accuracy}\\
    K = 100 & 1.53 & 89.55 \\
    K = 250 & 1.76 & 93.03 \\
    K = 500 & 2.30 & 94.08 \\
\end{tabular}
\caption{Case study results on our business conversation data by varying the value to retrieve the top K candidates. Here, ``Avg. Inf. Time'' refers to ``Average Inference Time in Seconds per utterance'', }
\label{tablecs}
\end{table}

\subsection{Case Study}

For the case study (see Table \ref{tablecs}), we conduct experiments with some additional values of $K$ for candidate retrieval to investigate its effect on accuracy and inference speed. For that purpose, in addition to the original value of $K$ = $250$ for the \textbf{BLINK + ES\textsubscript{MMQ}} model, we use the following values: $K$ = $100$ and $K$ = $500$. We find that even though reducing the value of $K$ to $100$ for candidate retrieval leads to a faster inference speed, the accuracy is decreased by 3.74\%. Moreover, increasing the value of $K$ to $500$ provides an opposite impact, as it improves the accuracy by 1.13\% but makes the candidate retrieval speed slower by taking more than 2 seconds per utterance. This trade-off implies that the retrieval value for $K$ can be tuned based on the requirement.

\subsection{Ablation Study}
To further investigate the effectiveness of our proposed approach of combining BLINK with Elasticsearch via leveraging MMQ for candidate retrieval, we do an ablation test. In our ablation test, we remove BLINK and only utilize the MMQ of Elasticsearch to retrieve the most relevant candidate. In this way, only one top matched candidate entity is retrieved from Elasticsearch. The result of our experiment is given in Table \ref{tableas}.

\begin{table}[t]
\centering
\begin{tabular}{c|c|c}
   \textbf{Model} &   \textbf{Avg. Inf. Time} & \textbf{Accuracy}\\
  
  {BLINK + ES\textsubscript{MMQ}} &  1.76 & 93.03 \\
   \textit{without} BLINK & 0.55 & 74.22 \\
\end{tabular}
\caption{Ablation test results on our business conversation data. Here, ``Avg. Inf. Time'' refers to ``Average Inference Time in Seconds per utterance'', }
\label{tableas}
\end{table}

From Table \ref{tableas}, we observe that even though removing BLINK led to a great improvement in terms of the inference speed, there is a significant drop in accuracy (by 20.22\%). This makes the model without BLINK inapplicable in production scenarios where the requirement is to ensure high accuracy. 

\section{Conclusion} In this paper, we introduce an efficient, scalable version of the BLINK model and extend it for entity linking on Wikidata. With extensive experiments, we show that our proposed system is usable for production environments within a limited budget setting since it significantly reduces memory requirements, computing resource usage, as well as the inference time while retaining high accuracy. We also effectively deploy our proposed entity linking system in a 10GB RAM machine without using any GPU for near real-time inference.  
In the future, we will investigate how to make our entity linking system more efficient such that it can give inference in real-time (e.g., within one second). Moreover, we will study how different BERT-based \cite{DBLP:journals/corr/distilbert, DBLP:conf/naacl/bert, liu2019roberta, lan2019albert} sentence similarity models \cite{tanda2019,laskar2020wslds, laskar-LREC,laskar2021domain} for candidate retrieval can impact the performance, while also exploring different techniques such as dimensionality reduction \cite{wang2016auto} to optimize the space used in Elasticsearch as well as the computing resource requirements.

\section{Ethics Statement}

The business phone conversational data used for entity linking experiments is annotated by the in-house Scientists for which the annotations were acquired for individual utterances. Whereas to annotate the conversation dataset to train our internal NER model, Appen was used (\url{https://appen.com/}) for data annotation and the annotators were provided with adequate compensation (above minimum wages). There is a data retention policy available for all users so that data will not be collected if the user is not consent to data collection. To protect user privacy, sensitive data such as personally identifiable information (e.g., credit card number, phone number) were removed while collecting the data. Since our model is doing classification to link the named entities to their corresponding entries in a publicly available knowledge base for information extraction, incorrect predictions will not cause any harm to the user besides an unsatisfactory experience. We also maintain the licensing requirements accordingly while using different tools, such as Wikidata, WikiMapper, PyWikiBot, Elasticsearch, HuggingFace, BLINK, etc. 

\section*{Acknowledgements}
We gratefully appreciate the reviewers for their excellent review comments that helped us to improve the quality of this paper.

\bibliography{anthology,custom}

\begin{thebibliography}{30}
\expandafter\ifx\csname natexlab\endcsname\relax\def\natexlab#1{#1}\fi

\bibitem[{Akbik et~al.(2019)Akbik, Bergmann, Blythe, Rasul, Schweter, and
  Vollgraf}]{DBLP:conf/naacl/AkbikBBRSV19}
Alan Akbik, Tanja Bergmann, Duncan Blythe, Kashif Rasul, Stefan Schweter, and
  Roland Vollgraf. 2019.
\newblock \href {https://doi.org/10.18653/v1/n19-4010} {{FLAIR:} an easy-to-use
  framework for state-of-the-art {NLP}}.
\newblock In \emph{Proceedings of the 2019 Conference of the North American
  Chapter of the Association for Computational Linguistics: Human Language
  Technologies, {NAACL-HLT} 2019, Minneapolis, MN, USA, June 2-7, 2019,
  Demonstrations}, pages 54--59. Association for Computational Linguistics.

\bibitem[{Akbik et~al.(2018)Akbik, Blythe, and
  Vollgraf}]{DBLP:conf/coling/AkbikBV18}
Alan Akbik, Duncan Blythe, and Roland Vollgraf. 2018.
\newblock \href {https://aclanthology.org/C18-1139/} {Contextual string
  embeddings for sequence labeling}.
\newblock In \emph{Proceedings of the 27th International Conference on
  Computational Linguistics, {COLING} 2018, Santa Fe, New Mexico, USA, August
  20-26, 2018}, pages 1638--1649. Association for Computational Linguistics.

\bibitem[{Auer et~al.(2007)Auer, Bizer, Kobilarov, Lehmann, Cyganiak, and
  Ives}]{auer2007dbpedia}
S{\"o}ren Auer, Christian Bizer, Georgi Kobilarov, Jens Lehmann, Richard
  Cyganiak, and Zachary Ives. 2007.
\newblock Dbpedia: A nucleus for a web of open data.
\newblock In \emph{The semantic web}, pages 722--735. Springer.

\bibitem[{Devlin et~al.(2019)Devlin, Chang, Lee, and
  Toutanova}]{DBLP:conf/naacl/bert}
Jacob Devlin, Ming{-}Wei Chang, Kenton Lee, and Kristina Toutanova. 2019.
\newblock \href {https://doi.org/10.18653/v1/n19-1423} {{BERT:} pre-training of
  deep bidirectional transformers for language understanding}.
\newblock In \emph{Proceedings of the 2019 Conference of the North American
  Chapter of the Association for Computational Linguistics: Human Language
  Technologies, {NAACL-HLT} 2019, Minneapolis, MN, USA, June 2-7, 2019, Volume
  1 (Long and Short Papers)}, pages 4171--4186. Association for Computational
  Linguistics.

\bibitem[{Fabian et~al.(2007)Fabian, Gjergji, Gerhard et~al.}]{fabian2007yago}
MS~Fabian, Kasneci Gjergji, WEIKUM Gerhard, et~al. 2007.
\newblock Yago: A core of semantic knowledge unifying wordnet and wikipedia.
\newblock In \emph{16th International World Wide Web Conference, WWW}, pages
  697--706.

\bibitem[{Gabrilovich et~al.(2013)Gabrilovich, Ringgaard, and
  Subramanya}]{wikiwned}
Evgeniy Gabrilovich, Michael Ringgaard, and Amarnag Subramanya. 2013.
\newblock Facc1: Freebase annotation of clueweb corpora, version 1 (release
  date 2013-06-26, format version 1, correction level 0).

\bibitem[{Garg et~al.(2019)Garg, Vu, and Moschitti}]{tanda2019}
Siddhant Garg, Thuy Vu, and Alessandro Moschitti. 2019.
\newblock Tanda: Transfer and adapt pre-trained transformer models for answer
  sentence selection.
\newblock \emph{arXiv preprint arXiv:1911.04118}.

\bibitem[{Guo and Barbosa(2018)}]{guo2018robust}
Zhaochen Guo and Denilson Barbosa. 2018.
\newblock Robust named entity disambiguation with random walks.
\newblock \emph{Semantic Web}, 9(4):459--479.

\bibitem[{Hoffart et~al.(2011)Hoffart, Yosef, Bordino, F{\"u}rstenau, Pinkal,
  Spaniol, Taneva, Thater, and Weikum}]{hoffart2011robust}
Johannes Hoffart, Mohamed~Amir Yosef, Ilaria Bordino, Hagen F{\"u}rstenau,
  Manfred Pinkal, Marc Spaniol, Bilyana Taneva, Stefan Thater, and Gerhard
  Weikum. 2011.
\newblock Robust disambiguation of named entities in text.
\newblock In \emph{Proceedings of the 2011 Conference on Empirical Methods in
  Natural Language Processing}, pages 782--792.

\bibitem[{Huang et~al.(2015)Huang, Xu, and Yu}]{DBLP:journals/corr/HuangXY15}
Zhiheng Huang, Wei Xu, and Kai Yu. 2015.
\newblock \href {http://arxiv.org/abs/1508.01991} {Bidirectional {LSTM-CRF}
  models for sequence tagging}.
\newblock \emph{CoRR}, abs/1508.01991.

\bibitem[{Johnson et~al.(2021)Johnson, Douze, and
  J{\'{e}}gou}]{DBLP:journals/tbd/faiss}
Jeff Johnson, Matthijs Douze, and Herv{\'{e}} J{\'{e}}gou. 2021.
\newblock \href {https://doi.org/10.1109/TBDATA.2019.2921572} {Billion-scale
  similarity search with gpus}.
\newblock \emph{{IEEE} Trans. Big Data}, 7(3):535--547.

\bibitem[{Lan et~al.(2019)Lan, Chen, Goodman, Gimpel, Sharma, and
  Soricut}]{lan2019albert}
Zhenzhong Lan, Mingda Chen, Sebastian Goodman, Kevin Gimpel, Piyush Sharma, and
  Radu Soricut. 2019.
\newblock Albert: A lite bert for self-supervised learning of language
  representations.
\newblock \emph{arXiv preprint arXiv:1909.11942}.

\bibitem[{Laskar et~al.(2021)Laskar, Hoque, and Huang}]{laskar2021domain}
Md~Tahmid~Rahman Laskar, Enamul Hoque, and Jimmy~Xiangji Huang. 2021.
\newblock Domain adaptation with pre-trained transformers for query focused
  abstractive text summarization.
\newblock \emph{arXiv preprint arXiv:2112.11670}.

\bibitem[{Laskar et~al.(2020{\natexlab{a}})Laskar, Hoque, and
  Huang}]{laskar2020wslds}
Md~Tahmid~Rahman Laskar, Enamul Hoque, and Xiangji Huang. 2020{\natexlab{a}}.
\newblock {WSL-DS}: Weakly supervised learning with distant supervision for
  query focused multi-document abstractive summarization.
\newblock In \emph{Proceedings of the 28th International Conference on
  Computational Linguistics}, pages 5647--5654.

\bibitem[{Laskar et~al.(2020{\natexlab{b}})Laskar, Huang, and
  Hoque}]{laskar-LREC}
Md~Tahmid~Rahman Laskar, Xiangji Huang, and Enamul Hoque. 2020{\natexlab{b}}.
\newblock Contextualized embeddings based transformer encoder for sentence
  similarity modeling in answer selection task.
\newblock In \emph{Proceedings of the 12th Language Resources and Evaluation
  Conference}, pages 5505--5514.

\bibitem[{Lin et~al.(2021)Lin, Gao, Yan, Moreno, and Ren}]{lin2021rockner}
Bill~Yuchen Lin, Wenyang Gao, Jun Yan, Ryan Moreno, and Xiang Ren. 2021.
\newblock Rockner: A simple method to create adversarial examples for
  evaluating the robustness of named entity recognition models.
\newblock In \emph{Proceedings of the 2021 Conference on Empirical Methods in
  Natural Language Processing}, pages 3728--3737.

\bibitem[{Liu et~al.(2019)Liu, Ott, Goyal, Du, Joshi, Chen, Levy, Lewis,
  Zettlemoyer, and Stoyanov}]{liu2019roberta}
Yinhan Liu, Myle Ott, Naman Goyal, Jingfei Du, Mandar Joshi, Danqi Chen, Omer
  Levy, Mike Lewis, Luke Zettlemoyer, and Veselin Stoyanov. 2019.
\newblock Roberta: A robustly optimized bert pretraining approach.
\newblock \emph{arXiv preprint arXiv:1907.11692}.

\bibitem[{M{\"o}ller et~al.(2021)M{\"o}ller, Lehmann, and
  Usbeck}]{moller2021survey}
Cedric M{\"o}ller, Jens Lehmann, and Ricardo Usbeck. 2021.
\newblock Survey on english entity linking on wikidata.
\newblock \emph{arXiv preprint arXiv:2112.01989}.

\bibitem[{Onoe and Durrett(2020{\natexlab{a}})}]{DBLP:conf/aaai/OnoeD20}
Yasumasa Onoe and Greg Durrett. 2020{\natexlab{a}}.
\newblock \href {https://aaai.org/ojs/index.php/AAAI/article/view/6380}
  {Fine-grained entity typing for domain independent entity linking}.
\newblock In \emph{The Thirty-Fourth {AAAI} Conference on Artificial
  Intelligence, {AAAI} 2020, The Thirty-Second Innovative Applications of
  Artificial Intelligence Conference, {IAAI} 2020, The Tenth {AAAI} Symposium
  on Educational Advances in Artificial Intelligence, {EAAI} 2020, New York,
  NY, USA, February 7-12, 2020}, pages 8576--8583. {AAAI} Press.

\bibitem[{Onoe and Durrett(2020{\natexlab{b}})}]{onoe2020interpretable}
Yasumasa Onoe and Greg Durrett. 2020{\natexlab{b}}.
\newblock Interpretable entity representations through large-scale typing.
\newblock In \emph{Proceedings of the 2020 Conference on Empirical Methods in
  Natural Language Processing: Findings}, pages 612--624.

\bibitem[{Peters et~al.(2018)Peters, Neumann, Iyyer, Gardner, Clark, Lee, and
  Zettlemoyer}]{peters-etal-2018-deep}
Matthew~E. Peters, Mark Neumann, Mohit Iyyer, Matt Gardner, Christopher Clark,
  Kenton Lee, and Luke Zettlemoyer. 2018.
\newblock \href {https://doi.org/10.18653/v1/N18-1202} {Deep contextualized
  word representations}.
\newblock In \emph{Proceedings of the 2018 Conference of the North {A}merican
  Chapter of the Association for Computational Linguistics: Human Language
  Technologies, Volume 1 (Long Papers)}, pages 2227--2237, New Orleans,
  Louisiana. Association for Computational Linguistics.

\bibitem[{Ravi et~al.(2021)Ravi, Singh, Mulang, Shekarpour, Hoffart, and
  Lehmann}]{DBLP:conf/eacl/cholan}
Manoj Prabhakar~Kannan Ravi, Kuldeep Singh, Isaiah~Onando Mulang, Saeedeh
  Shekarpour, Johannes Hoffart, and Jens Lehmann. 2021.
\newblock \href {https://aclanthology.org/2021.eacl-main.40/} {{CHOLAN:} {A}
  modular approach for neural entity linking on wikipedia and wikidata}.
\newblock In \emph{Proceedings of the 16th Conference of the European Chapter
  of the Association for Computational Linguistics: Main Volume, {EACL} 2021,
  Online, April 19 - 23, 2021}, pages 504--514. Association for Computational
  Linguistics.

\bibitem[{Sakor et~al.(2020)Sakor, Singh, Patel, and
  Vidal}]{DBLP:conf/cikm/falcon}
Ahmad Sakor, Kuldeep Singh, Anery Patel, and Maria{-}Esther Vidal. 2020.
\newblock \href {https://doi.org/10.1145/3340531.3412777} {Falcon 2.0: An
  entity and relation linking tool over wikidata}.
\newblock In \emph{{CIKM} '20: The 29th {ACM} International Conference on
  Information and Knowledge Management, Virtual Event, Ireland, October 19-23,
  2020}, pages 3141--3148. {ACM}.

\bibitem[{Sanh et~al.(2019)Sanh, Debut, Chaumond, and
  Wolf}]{DBLP:journals/corr/distilbert}
Victor Sanh, Lysandre Debut, Julien Chaumond, and Thomas Wolf. 2019.
\newblock \href {http://arxiv.org/abs/1910.01108} {Distilbert, a distilled
  version of {BERT:} smaller, faster, cheaper and lighter}.
\newblock \emph{CoRR}, abs/1910.01108.

\bibitem[{Shen et~al.(2014)Shen, Wang, and Han}]{shen2014entity}
Wei Shen, Jianyong Wang, and Jiawei Han. 2014.
\newblock Entity linking with a knowledge base: Issues, techniques, and
  solutions.
\newblock \emph{IEEE Transactions on Knowledge and Data Engineering},
  27(2):443--460.

\bibitem[{Vaswani et~al.(2017)Vaswani, Shazeer, Parmar, Uszkoreit, Jones,
  Gomez, Kaiser, and Polosukhin}]{DBLP:conf/nips/VaswaniSPUJGKP17}
Ashish Vaswani, Noam Shazeer, Niki Parmar, Jakob Uszkoreit, Llion Jones,
  Aidan~N. Gomez, Lukasz Kaiser, and Illia Polosukhin. 2017.
\newblock \href
  {https://proceedings.neurips.cc/paper/2017/hash/3f5ee243547dee91fbd053c1c4a845aa-Abstract.html}
  {Attention is all you need}.
\newblock In \emph{Advances in Neural Information Processing Systems 30: Annual
  Conference on Neural Information Processing Systems 2017, December 4-9, 2017,
  Long Beach, CA, {USA}}, pages 5998--6008.

\bibitem[{Wang et~al.(2016)Wang, Yao, and Zhao}]{wang2016auto}
Yasi Wang, Hongxun Yao, and Sicheng Zhao. 2016.
\newblock Auto-encoder based dimensionality reduction.
\newblock \emph{Neurocomputing}, 184:232--242.

\bibitem[{Wolf et~al.(2020)Wolf, Chaumond, Debut, Sanh, Delangue, Moi, Cistac,
  Funtowicz, Davison, Shleifer et~al.}]{wolf2020transformers}
Thomas Wolf, Julien Chaumond, Lysandre Debut, Victor Sanh, Clement Delangue,
  Anthony Moi, Pierric Cistac, Morgan Funtowicz, Joe Davison, Sam Shleifer,
  et~al. 2020.
\newblock Transformers: State-of-the-art natural language processing.
\newblock In \emph{Proceedings of the 2020 Conference on Empirical Methods in
  Natural Language Processing: System Demonstrations}, pages 38--45.

\bibitem[{Wu et~al.(2020)Wu, Petroni, Josifoski, Riedel, and
  Zettlemoyer}]{DBLP:conf/emnlp/blink}
Ledell Wu, Fabio Petroni, Martin Josifoski, Sebastian Riedel, and Luke
  Zettlemoyer. 2020.
\newblock \href {https://doi.org/10.18653/v1/2020.emnlp-main.519} {Scalable
  zero-shot entity linking with dense entity retrieval}.
\newblock In \emph{Proceedings of the 2020 Conference on Empirical Methods in
  Natural Language Processing, {EMNLP} 2020, Online, November 16-20, 2020},
  pages 6397--6407. Association for Computational Linguistics.

\bibitem[{Yamada et~al.(2020)Yamada, Asai, Shindo, Takeda, and
  Matsumoto}]{DBLP:conf/emnlp/YamadaASTM20}
Ikuya Yamada, Akari Asai, Hiroyuki Shindo, Hideaki Takeda, and Yuji Matsumoto.
  2020.
\newblock \href {https://doi.org/10.18653/v1/2020.emnlp-main.523} {{LUKE:} deep
  contextualized entity representations with entity-aware self-attention}.
\newblock In \emph{Proceedings of the 2020 Conference on Empirical Methods in
  Natural Language Processing, {EMNLP} 2020, Online, November 16-20, 2020},
  pages 6442--6454. Association for Computational Linguistics.

\end{thebibliography}
\bibliographystyle{acl_natbib}

\appendix


\end{document}